# Clustering by Low-Rank Doubly Stochastic Matrix Decomposition

**Zhirong Yang**                                                                                          ZHIRONG.YANG@AALTO.FI
Department of Information and Computer Science, Aalto University, 00076, Finland

**Erkki Oja**                                                                                                ERKKI.OJA@AALTO.FI
Department of Information and Computer Science, Aalto University, 00076, Finland

## Abstract

Clustering analysis by nonnegative low-rank approximations has achieved remarkable progress in the past decade. However, most approximation approaches in this direction are still restricted to matrix factorization. We propose a new low-rank learning method to improve the clustering performance, which is beyond matrix factorization. The approximation is based on a two-step bipartite random walk through virtual cluster nodes, where the approximation is formed by only cluster assigning probabilities. Minimizing the approximation error measured by Kullback-Leibler divergence is equivalent to maximizing the likelihood of a discriminative model, which endows our method with a solid probabilistic interpretation. The optimization is implemented by a relaxed Majorization-Minimization algorithm that is advantageous in finding good local minima. Furthermore, we point out that the regularized algorithm with Dirichlet prior only serves as initialization. Experimental results show that the new method has strong performance in clustering purity for various datasets, especially for large-scale manifold data.

## 1. Introduction

Cluster analysis assigns a set of objects into groups so that the objects in the same cluster are more similar to each other than to those in other clusters. Optimization of most clustering objectives is NP-hard and relaxation to "soft" clustering is often required. A non-negativity constraint, together with various low-rank matrix approximation objectives, has widely been used for the relaxation purpose in the past decade.

The most popular nonnegative low-rank approximation method is *Nonnegative Matrix Factorization* (NMF). It finds a matrix that approximates the similarities and can be factorized into several nonnegative low-rank matrices. NMF was originally applied to vectorial data, where Ding et al. (2010) have shown that NMF is equivalent to the classical $k$-means method. Later NMF was applied to the (weighted) graph given by the pairwise similarities. For example, Ding et al. (2008) presented Nonnegative Spectral Cuts by using a multiplicative algorithm; Arora et al. (2011) proposed Left Stochastic Decomposition that approximates a similarity matrix based on Euclidean distance and a left-stochastic matrix. Another stream in the same direction is topic modeling. Hofmann (1999) gave a generative model in *Probabilistic Latent Semantic Indexing* (PLSI) for counting data, which is essentially equivalent to NMF using Kullback-Leibler (KL) divergence and Tri-factorizations. Bayesian treatment of PLSI by using Dirichlet prior was later introduced by Blei et al. (2001). Symmetric PLSI with the same Bayesian treatment is called *Interaction Component Model* (ICM) (Sinkkonen et al., 2008).

Despite remarkable progress, the above relaxation approaches are still not fully satisfactory in all of the following requirements that affect the clustering performance using nonnegative low-rank approximation: (1) *approximation error measure* that takes into account sparse similarities, (2) *decomposition form* of the approximating matrix, where the decomposing matrices should contain just enough parameters for clustering but not more, and (3) *normalization* of the approximating matrix, which ensures relatively balanced clusters and equal contribution of each data sample. Lacking one or more of these dimensions can severely affect clustering performance.

In this paper we present a new nonnegative low-rank approximation method for clustering, which satisfies all of the above three requirements. First, because datasets often lie in curved manifolds such that only similarities in a small neighborhood are reliable, we adopt KL-divergence to handle the resulting sparsity. Second, different from PLSI, we enforce an equal contribution of every data sample and then directly construct the decomposition over the probabilities from samples to clusters. Third, these probabilities form the only decomposing matrix to be learned in our approach and directly gives the answer for probabilistic clustering. Furthermore, our decomposition method leads to a doubly-stochastic approximating matrix, which was shown to be desired for balanced graph cuts (Zass & Shashua, 2006). We name our new method DCD because it is based on Data-Cluster-Data random walks.

In order to solve the DCD learning objective, we propose a novel relaxed Majorization-Minimization algorithm to handle the new matrix decomposition type. Our relaxation strategy works robustly in finding sastisfactory local optimizers under the stochasticity constraint. Furthermore, we argue that complexity control such as Bayesian priors only provides initialization for the new algorithm. This eliminates the problem of hyperparameter selection in the prior.

Empirical comparison with NMF and other graph-based clustering approaches demonstrates that our method can achieve the best or nearly the best clustering purity in all tasks. For some datasets, the new method significantly improves the state-of-the-art.

After this introductory part, we present the new method in Section 2, including its learning objective, probabilistic model, optimization and initialization techniques. In Section 3, we point out the connections and differences between our method and other recent related work. Experimental settings and results are given in Section 4. Finally we conclude the paper and discuss some future work in Section 5.

## 2. Clustering by DCD

Suppose the similarities between $n$ data samples are precomputed and given in a nonnegative symmetric matrix $A$. This matrix can be seen as (weighted) affinity of an undirected similarity graph where each node corresponds to a data sample (data node). A clustering analysis algorithm takes such input and divides the data nodes into $r$ disjoint subsets. In probabilistic clustering analysis, we want to find $P(k|i)$, the probability of assigning the $i$th sample to the $k$th cluster, where $i = 1, \ldots, n$ and $k = 1, \ldots, r$. In the following, $i, j$ and $v$ stand for data sample (node) indices while $k$ and $l$ stand for cluster indices.

### 2.1. Learning objective

Some of our work was inspired by the AnchorGraph (Liu et al., 2010) which was used in large approximative graph construction based on a two-step random walk between data nodes through a set of anchor nodes. Note that AnchorGraph is not a clustering method.

If we augment the input similarity graph by $r$ *cluster nodes*, the cluster assigning probabilities can be seen as single-step random walk probabilities from data nodes to the augmented cluster nodes. Without preference to any particular samples, we impose uniform prior $P(i) = 1/n$ over the data nodes. By this prior, the reversed random walk probabilities can then be calculated by the Bayes formula

$$P(i|k) = \frac{P(k|i)P(i)}{\sum_v P(k|v)P(v)} = \frac{P(k|i)}{\sum_v P(k|v)}. \quad (1)$$

Consider next the probability of two-step random walks from $i$th data node to $j$th data node via all cluster nodes (*DCD random walk*):

$$P(i|j) = \sum_k P(i|k)P(k|j) = \sum_k \frac{P(k|i)P(k|j)}{\sum_v P(k|v)}. \quad (2)$$

This probability defines another similarity between two data nodes, $\widehat{A}_{ij} = P(i|j)$, with respect to cluster nodes. Note that this matrix has rank at most equal to $r$. The learning target is now to find a good approximation between the input similarities and the DCD random walk probabilities:

$$A \approx \widehat{A}. \quad (3)$$

AnchorGraph does not provide any error measure for the above approximation. A conventional choice in NMF is the squared Euclidean distance, which employs the underlying assumption that the noise is additive and Gaussian.

In real-world clustering tasks for multivariate datasets, data points often lie in a curved manifold. Consequently, similarities based on Euclidean distances are reliable only in a small neighborhood. Such locality causes high sparsity in the input similarity matrix. Sparsity also commonly exists for real-world network data. Because of the sparsity, Euclidean distance is improper for the approximation in Eq. (3), because additive Gaussian noise should lead to a dense

Clustering by Low-Rank Doubly Stochastic Matrix Decompositionobserved graph. In contrast, (generalized) Kullback-Leibler divergence is more suitable for the approximation. The underlying Poisson noise characterizes rare occurrences that are present in our sparse input. We can now formulate our learning objective as the following optimization problem:

$$\min_{W \geq 0} \ D_{\mathrm{KL}}(A||\widehat{A}) = \sum_{ij} \left( A_{ij} \log \frac{A_{ij}}{\widehat{A}_{ij}} - A_{ij} + \widehat{A}_{ij} \right) \quad (4)$$

$$\text{s.t.} \ \sum_k W_{ik} = 1, \ i = 1, \ldots, n, \quad (5)$$

where we write $W_{ik} = P(k|i)$ for convenience and thus

$$\widehat{A}_{ij} = \sum_k \frac{W_{ik} W_{jk}}{\sum_v W_{vk}}. \quad (6)$$

Note that $\widehat{A}$ is symmetric as it is easy to verify that $P(i|j) = P(j|i)$. Therefore, $\widehat{A}$ is also doubly stochastic because it is left-stochastic by probability definition.

### 2.2. Probabilistic model

The optimization objective has an analogous statistical model with the PLSI. Dropping the constant terms from $D_{\mathrm{KL}}(A||\widehat{A})$, the objective is equivalent to maximizing

$$\sum_{ij} A_{ij} \log \widehat{A}_{ij}. \quad (7)$$

This can be identified as the log-likelihood of the following generative model if $A_{ij}$ are integers: for $t = 1, \ldots, T$, add one to entry $(i, j) \sim$ Multinomial $\left(\frac{1}{n}\widehat{A}, 1\right)$, whose likelihood is given by

$$p(A) = \prod_{t=1}^T \frac{1}{n} \widehat{A}_{ij} = \prod_{ij} \left( \frac{1}{n} \widehat{A}_{ij} \right)^{A_{ij}},$$

where $T = \sum_{ij} A_{ij}$.

The above model simply uses uniform prior on rows of $W$. It does not prevent from using informative priors or complexity control. A natural choice for probabilities is the Dirichlet distribution ($\alpha > 0$)

$$p(W_{i1}, \ldots, W_{ir} | \alpha) = \frac{\Gamma(r\alpha)}{[\Gamma(\alpha)]^r} \prod_{k=1}^r W_{ik}^{\alpha-1}, \quad (8)$$

which is also the conjugate prior of multinomial distribution. The Dirichlet prior reduces to be uniform when $\alpha = 1$.

Although it is possible to construct a multi-level graphical model similar to the Dirichlet process topic model, we emphasize that the smallest approximation error (or perplexity) is our final goal. Dirichlet prior is used only in order to ease the optimization. Therefore we do not employ more complex generative models; see Section 2.4 for more discussion.

### 2.3. Optimization

The optimization problem with Dirichlet prior on $W$ is equivalent to minimizing

$$\mathcal{J}(W) = -\sum_{ij} A_{ij} \log \widehat{A}_{ij} - (\alpha - 1) \sum_{ik} \log W_{ik} \quad (9)$$

There are two ways to handle the constraint Eq. (5). First, one can develop the multiplicative algorithm by the procedure proposed by Yang & Oja (2011) by neglecting the stochasticity constraint, and then normalize the rows of $W$ after each update. However, the optimization by this way easily gets stuck in poor local minima in practice.

Here we employ a relaxing strategy to handle the constraint. We first introduce Lagrangian multipliers for the constraints:

$$\mathcal{L}(W, \lambda) = \mathcal{J}(W) + \sum_i \lambda_i \left( \sum_k W_{ik} - 1 \right). \quad (10)$$

Unlike traditional constrained optimization that solves the fixed-point equations, we employ a heuristic to find the multipliers $\lambda$. Denote $\nabla = \nabla^+ - \nabla^-$ the gradient of $\mathcal{J}$ with respect to $W$, where $\nabla^+$ and $\nabla^-$ are the positive and (unsigned) negative parts, respectively. This suggests a fixed-point update rule for $W$:

$$W'_{ik} = W_{ik} \frac{\nabla^-_{ik} - \lambda_i}{\nabla^+_{ik}}. \quad (11)$$

Imposing $\sum_k W'_{ik} = 1$, we obtain

$$\lambda_i = \frac{b_i - 1}{a_i}, \quad (12)$$

where $a_i$ and $b_i$ are given in Algorithm 1. Next we show that the augmented objective Eq. (10) decreases after each iteration with the above $\lambda$.

**Theorem 1.** *Denote $W^{new}$ the updated matrix after each iteration. It holds that $\mathcal{L}(W^{new}, \lambda) \leq \mathcal{L}(W, \lambda)$ with $\lambda_i = (b_i - 1)/a_i$.*

*Proof.* The algorithm construction mainly follows the Majorization-Minimization procedure (see e.g. Yang & Oja, 2011). We use $W$ and $\widetilde{W}$ to distinguish the current estimate and the variable, respectively.

**Algorithm 1** Relaxed MM Algorithm for DCD
**Input:** similarity matrix $A$, number of clusters $r$, nonnegative initial guess of $W$.
**repeat**
$$Z_{ij} = \left(\sum_k \frac{W_{ik}W_{jk}}{\sum_v W_{vk}}\right)^{-1} A_{ij}$$
$$s_k = \sum_v W_{vk}$$
$$\nabla_{ik}^- = 2(ZW)_{ik} s_k^{-1} + \alpha W_{ik}^{-1}$$
$$\nabla_{ik}^+ = (W^T Z W)_{kk} s_k^{-2} + W_{ik}^{-1}$$
$$a_i = \sum_l \frac{W_{il}}{\nabla_{il}^+}, \quad b_i = \sum_l W_{il} \frac{\nabla_{il}^-}{\nabla_{il}^+}$$
$$W_{ik} \leftarrow W_{ik} \frac{\nabla_{ik}^- a_i + 1}{\nabla_{ik}^+ a_i + b_i}$$
**until** $W$ is unchanged
**Output:** cluster assigning probabilities $W$.

**(Majorization)**

Let $\phi_{ijk} = \frac{W_{ik}W_{jk}}{\sum_v W_{vk}} \left(\sum_l \frac{W_{il}W_{jl}}{\sum_v W_{vl}}\right)^{-1}$.

$$\mathcal{L}(\widetilde{W})$$
$$\leq -\sum_{ijk} A_{ij}\phi_{ijk}\left[\log \widetilde{W}_{ik} + \log \widetilde{W}_{jk} - \log \sum_v \widetilde{W}_{vk}\right]$$
$$-(\alpha-1)\sum_{ik}\log \widetilde{W}_{ik} + \sum_{ik}\lambda_i W_{ik} + C_1$$
$$\leq -\sum_{ijk} A_{ij}\phi_{ijk}\left[\log \widetilde{W}_{ik} + \log \widetilde{W}_{jk} - \frac{\sum_v \widetilde{W}_{vk}}{\sum_v W_{vk}}\right]$$
$$-(\alpha-1)\sum_{ik}\log \widetilde{W}_{ik} + \sum_{ik}\lambda_i W_{ik} + C_2$$
$$\leq -\sum_{ijk} A_{ij}\phi_{ijk}\left[\log \widetilde{W}_{ik} + \log \widetilde{W}_{jk} - \frac{\sum_v \widetilde{W}_{vk}}{\sum_v W_{vk}}\right]$$
$$-(\alpha-1)\sum_{ik}\log \widetilde{W}_{ik} + \sum_{ik}\lambda_i W_{ik}$$
$$+ \sum_{ik}\left(\frac{1}{a_i} + \frac{1}{W_{ik}}\right) W_{ik}\left(\frac{\widetilde{W}_{ik}}{W_{ik}} - \log \frac{\widetilde{W}_{ik}}{W_{ik}} - 1\right) + C_2$$
$$\equiv G(\widetilde{W}, W),$$

where $C_1$ and $C_2$ are constants irrelevant to the variable $\widetilde{W}$. The first two inequalities follow the CCCP majorization (Yang & Oja, 2011) using the convexity and concavity of $-\log()$ and $\log()$, respectively. The third inequality is called "moving term" technique used in multiplicative updates (Yang & Oja, 2010).

It adds the same constant $\frac{1}{a_i} + \frac{1}{W_{ik}}$ to both numerator and denominator in order to guarantee that the updated matrix entries are positive, which is implemented by using a further upper-bound of zero. All the above upper bounds are tight at $\widetilde{W} = W$, i.e. $G(W, W) = \mathcal{J}(W)$.

**(Minimization)**

$$\frac{\partial G}{\partial \widetilde{W}_{ik}} = \nabla_{ik}^+ - \frac{1}{W_{ik}} - \frac{W_{ik}}{\widetilde{W}_{ik}}\left(\nabla_{ik}^- - \frac{1}{W_{ik}}\right)$$
$$+ \lambda_i + \left(\frac{1}{a_i} + \frac{1}{W_{ik}}\right) W_{ik}\left(\frac{1}{W_{ik}} - \frac{1}{\widetilde{W}_{ik}}\right)$$
$$= -\frac{W_{ik}}{\widetilde{W}_{ik}}\left(\nabla_{ik}^- + \frac{1}{a_i}\right) + \left(\nabla_{ik}^+ + \frac{b_i}{a_i}\right).$$

Setting the gradient to zero gives

$$W_{ik}^{\text{new}} = W_{ik} \frac{\nabla_{ik}^- + \frac{1}{a_i}}{\nabla_{ik}^+ + \frac{b_i}{a_i}} \quad (13)$$

Multiplying both numerator and denominator by $a_i$ gives the last update rule in Algorithm 1. Therefore, $\mathcal{L}(W^{\text{new}}, \lambda) \leq G(W^{\text{new}}, W) \leq \mathcal{L}(W, \lambda)$. $\square$

Algorithm 1 jointly minimizes the approximation error and drives the rows of $W$ towards the probability simplex. The Lagrangian multipliers are automatically selected by the algorithm, without extra human tuning labor. The quantities $b_i$ are the row sums of the unconstrained multiplicative learning result, while the quantities $a_i$ balance between the gradient learning force and the probability simplex attraction. Besides convenience, we find that this relaxation strategy works more robustly than the brute-force normalization after each iteration.

### 2.4. Initialization

The optimization problems of many clustering analysis methods, including ours, are non-convex. Usually finding the global optimum is very expensive or even NP-hard. When local optimizers are used, the optimization trajectory can easily get stuck in poor local optima if the algorithm starts from an arbitrary random guess. Proper initialization is thus needed to achieve satisfactory performance.

The cost of the initialization should be much cheaper than the main algorithm. There are two popular choices: *k-means* and *Normalized Cut* (Ncut). The first one can only be applied to vectorial data and could be slow for large-scale high-dimensional data. Here we employ the second initialization method. While the original Ncut is NP-hard, the relaxed Ncut problem can be efficiently solved via spectral methods (Shi & Malik, 2000). Furthermore, it is particularly suitable for sparse graph input, which is our focus in this paper.

Besides Ncut, we emphasize that *the minimal approximation error is our sole learning objective and all regularized versions, e.g. with different Dirichlet priors, only serve as initialization.* This is because clustering analysis, unlike supervised learning problems, does not need to provide inference for unseen data. That is, *the complexity control such as Bayesian priors is not meant for better generalization performance, but for better-shaped space to facilitate optimization.* In this sense, we can use the results of diverse regularized versions or even other clustering algorithms as starting guesses, and pick the one with the smallest approximation error among multiple runs.

In implementation, we first convert an initialization clustering result to an $n \times r$ binary indicator matrix, and then add a small positive perturbation to all entries. Next, the perturbed matrix is fed to our optimization algorithm (with $\alpha = 1$ in Algorithm 1).

## 3. Related Work

Our method intersects with several other machine learning approaches. Here we discuss some of these directions, pinpointing the connections and our new contributions.

### 3.1. Topic Model

A topic model is a type of statistical model for discovering the abstract "topics" that occur in a collection of documents. An early topic model was PLSI (Hofmann, 1999) which maximizes the following log-likelihood for symmetric input $A$:

$$\sum_{ij} A_{ij} \log \sum_k P(k)P(i|k)P(j|k). \qquad (14)$$

One can see that PLSI has similar form as Eq. (7). Both objectives can be equivalently expressed by nonnegative low-rank approximation using KL-divergence.

The major difference is the decomposition form of the approximating matrix. There are two ways to model the hierarchy between latent variables and the observed ones. Topic model uses the pure generative way while our method employs the discriminative way. PLSI gives the clustering results indirectly. One should apply Bayes formula to evaluate $P(k|i)$ using $P(i|k)$ and $P(k)$. There are $n \times r - 1$ free parameters to be learned in the latter two quantities. In contrast, our method directly learns the cluster assigning probabilities $P(k|i)$ which contains only $n \times (r-1)$ free parameters. This difference can be large when there are only a few clusters (e.g. $r = 2$ or $r = 3$).

It is known that the performance of PLSI can be improved by using Bayesian non-parametric modeling. Bayesian treatment for the symmetric version of PLSI leads to Interaction Component Model (Sinkkonen et al., 2008). It associates Dirichlet priors to the PLSI factorizing matrices and then makes use of the conjugacy between Dirichlet and multinomial to derive collapsed Gibbs sampling or variational optimization methods.

An open problem of Bayesian methods is how to determine the hyperparameters that control the priors. Asuncion et al. (2009) found that wrongly chosen parameters can lead to only mediocre or even poor performance. The automatic hyperparameters updating method proposed by Minka (2000) does not necessarily lead to good solution in terms of perplexity (Asuncion et al., 2009) or clustering purity in our experiments (see Section 4). Hofmann (1999); Asuncion et al. (2009) suggested to select the hyperparameters using the smallest approximation error for some heldout matrix entries, which is however more costly and might weaken or even break the cluster structure.

By contrast, there is no such prior hyperparameter selection problem in our method. The algorithms using various priors only play their role in the initialization. Among the runs with different starting points, we simply select the one with the smallest approximation error.

### 3.2. Nonnegative Matrix Factorization

Nonnegative Matrix Factorization is one of the earliest methods for relaxing clustering problems by nonnegative low-rank approximation (see e.g. Xu et al., 2003). The research on NMF also opened the door for multiplicative majorization-minimization algorithms for optimization over nonnegative matrices. In the original NMF, an input nonnegative matrix $X$ is approximated by a product of two low-rank matrices $W$ and $H$. Later researchers found that more constraints or normalizations should be imposed on the factorizing matrices to achieve desired performance.

Orthogonality is a popular choice (see e.g. Ding et al., 2006) for highly sparse factorizing matrices, especially the cluster indicator matrix. However, the orthogonality constraint seems exclusive of other constraints or priors. In practice, the orthogonality favors Euclidean distance as the approximation error measure for simple update rules, which is against our requirement for sparse graph input.

Stochasticity seems more natural for relaxing hard clustering to probabilities. Recently Arora et al. (2011) proposed a symmetric NMF using left-stochastic factorizing matrices called LSD. Their method also directly learns the cluster assigning probabilities. However, LSD is also restricted to Euclidean distance.

Our method has two major differences from LSD. First, we use Kullback-Leibler divergence which is more suitable for sparse graph input or curved manifold data. This also enables us to make use of the Dirichlet and multinomial conjugacy pair. Second, our decomposition has good interpretation in terms of a random walk. Furthermore, imbalanced clustering is implicitly penalized because of the denominator in Eq. (6).

### 3.3. AnchorGraph

DCD uses the same matrix decomposition as Anchor-Graph. However, there are several major differences between the two methods. First of all, AnchorGraph is not made for clustering, but for constructing the graph input. AnchorGraph has no learning objective that captures the global structure of data such as clusters. Each row of the decomposing matrix in AnchorGraph is learned individually and only encodes the local information. There is no learning over the decomposing matrix as a whole. Furthermore, anchors are either selected from data samples or pre-learned by e.g. $k$-means. By contrast, cluster nodes in our formulation are virtual. They are not vectors and need no physical storage.

## 4. Experiments

### 4.1. Compared methods

We have compared our method with eight other clustering algorithms that can take a symmetric nonnegative sparse matrix as input. The compared algorithms range from classical to state-of-the-art methods with various principles: graph cuts including Normalized Cut (Ncut) (Shi & Malik, 2000), Nonnegative Spectral Cut (NSC) (Ding et al., 2008), and 1-Spectral ratio Cheeger cut (1-Spec) (Hein & Bühler, 2010); nonnegative matrix factorization including Projective NMF (PNMF) (Yang & Oja, 2010), Symmetric 3-Factor Orthogonal NMF (ONMF) (Ding et al., 2006), and Left-Stochastic Decomposition (LSD) (Arora et al., 2011); topic models including Probabilistic Latent Semantic Indexing (PLSI) (Hofmann, 1999) and Interaction Component Model (ICM) (Sinkkonen et al., 2008).

Table 1. Statistics of selected datasets.

| DATASET | #SAMPLES | #CLASSES |
|---|---|---|
| AMAZON | 96 | 2 |
| IRIS | 150 | 3 |
| VOTES | 435 | 2 |
| ORL | 400 | 40 |
| PIE | 1166 | 53 |
| YALEB | 1292 | 38 |
| COIL20 | 1440 | 20 |
| ISOLET | 1559 | 26 |
| MFEAT | 2000 | 10 |
| WEBKB4 | 4196 | 4 |
| 7SECTORS | 4556 | 7 |
| USPS | 9298 | 10 |
| PENDIGITS | 10992 | 10 |
| LETRECO | 20000 | 26 |
| MNIST | 70000 | 10 |

The detailed settings of the compared methods are as follows. We implemented NSC, PNMF, ONMF, LSD, PLSI, and DCD using multiplicative updates. For these methods, we ran their update rules for 10,000 iterations to ensure that all algorithms have sufficiently converged. We used the default setting for 1-Spec. ICM uses collapsed Gibbs sampling, where each round of the sampling sweeps the graph once. We ran the ICM sampling for 100,000 rounds to ensure that the MCMC burn-in is converged (it took about one day for the largest dataset). The hyperparameters in ICM are automatically adjusted by using Minka's method (Minka, 2000).

Despite mediocre results, Ncut runs very fast and gives pretty stable outputs. We thus used it for initialization. After getting the Ncut cluster indicator matrix, we add 0.2 to all entries and feed it as starting point for other methods, which is a common initialization setting for NMF methods. The other three initialization points for our method are provided by Ncut followed by DCD using three different Dirichlet priors ($\alpha = 1.2$, $\alpha = 2$, and $\alpha = 5$). The clustering result of our method is reported by the run with the smallest approximation error, see Eq. (4).

### 4.2. Datasets

The performance of clustering methods were evaluated using real-world datasets. In particular, we focus on data that lie in a curved manifold. We thus selected 15 such datasets which are publicly available from a variety of domains. The data sources are given in the supplemental document.

The statistics of the selected datasets are summarized in Table 1. In brief, *Amazon* are book similarities according to amazon.com buying records; *Votes* are voting records in US congress by two different parties; *ORL, PIE, YaleB* are face images collected under different conditions; *COIL20* are small toy images; *Isolet* and *LegReco* are handwritten English letter images; *Webkb4* and *7sectors* are text document collections; *Mfeat, USPS, PenDigits, MNIST* are handwritten digit images.

Table 2. Clustering purities for the compared methods on various data sets.

| Dataset | Ncut | PNMF | NSC | ONMF | PLSI | LSD | 1-Spec | ICM | DCD |
|---|---|---|---|---|---|---|---|---|---|
| Amazon | 0.63 | 0.76 | 0.63 | 0.63 | 0.63 | 0.68 | 0.63 | 0.63 | **0.78** |
| Iris | 0.90 | 0.93 | 0.90 | 0.33 | 0.91 | **0.97** | 0.91 | **0.97** | **0.97** |
| Votes | 0.72 | 0.72 | 0.72 | **0.73** | **0.73** | 0.72 | 0.72 | **0.73** | **0.73** |
| ORL | 0.81 | 0.82 | 0.82 | 0.03 | **0.83** | 0.81 | 0.80 | 0.20 | **0.83** |
| PIE | 0.67 | 0.66 | 0.68 | 0.02 | 0.68 | **0.69** | 0.64 | 0.12 | 0.68 |
| YaleB | 0.45 | 0.43 | 0.46 | 0.03 | **0.51** | 0.45 | 0.39 | 0.10 | **0.51** |
| Coil20 | 0.81 | 0.71 | **0.82** | 0.05 | **0.82** | 0.78 | 0.75 | 0.63 | 0.81 |
| Isolet | 0.57 | 0.55 | 0.56 | 0.04 | **0.58** | 0.57 | 0.57 | 0.36 | **0.58** |
| Mfeat | 0.75 | 0.77 | 0.79 | 0.10 | 0.77 | 0.78 | **0.80** | 0.69 | 0.78 |
| Webkb4 | 0.54 | 0.41 | 0.54 | 0.40 | 0.59 | **0.62** | 0.40 | 0.49 | **0.62** |
| 7sectors | 0.25 | 0.29 | 0.25 | 0.24 | 0.37 | 0.35 | 0.25 | 0.38 | **0.41** |
| USPS | 0.74 | 0.75 | 0.74 | 0.77 | 0.73 | 0.79 | 0.74 | 0.60 | **0.81** |
| PenDigits | 0.80 | 0.78 | 0.80 | 0.10 | 0.80 | 0.86 | 0.80 | 0.52 | **0.89** |
| LetReco | 0.24 | 0.25 | 0.23 | 0.04 | 0.28 | 0.29 | 0.18 | 0.21 | **0.32** |
| MNIST | 0.77 | 0.74 | 0.79 | 0.11 | 0.79 | 0.76 | 0.88 | 0.95 | **0.97** |

We preprocessed the above datasets to produce similarity graph input except *Amazon* which is already in sparse graph format. We extracted the scattering features (Mallat, 2012) for image data except *Isolet* and *Mfeat* which have their own feature representation. We used Tf-Idf features for text documents.

After feature extraction, we constructed a *K-Nearest-Neighbor* (KNN) graph for each dataset. We set $K = 5$ for the six smallest datasets (except *Amazon*) and $K = 10$ for the other datasets. We then symmetrized and binarized the KNN graph $B$ to obtain the input similarities $A$ (i.e. $A_{ij} = 1$ if $B_{ij} = 1$ or $B_{ji} = 1$, and $A_{ij} = 0$ otherwise).

### 4.3. Results

Clustering performance of the compared methods is evaluated by *clustering purity*

$$\text{purity} = \frac{1}{n} \sum_{k=1}^{r} \max_{1 \leq l \leq r} n_k^l \qquad (15)$$

where $n_k^l$ is the number of data samples in the cluster $k$ that belong to ground-truth class $l$. A larger purity in general corresponds to better clustering result. The clustering purities for the compared methods are given in Table 2.

Our method has strong performance in terms of clustering purity. DCD wins 12 out of 15 selected datasets. Even for the other three datasets, DCD is the first or second runner-up, with purities tied with or very close to the winner.

The new method is particularly more advantageous for large datasets. Note that the datasets in Table 2 are ordered by their sizes. We can see that there are some other winners or joint winners for smaller datasets, for example, LSD for the *PIE* faces or 1-Spec for the *Mfeat* digits. PLSI performs quite similarly with DCD for these small clustering tasks. However, DCD demonstrates clear win over the other methods for the five largest datasets.

DCD has remarkable performance for the largest dataset *MNIST*. In this case, clustering as unsupervised learning by using our method has even achieved classification accuracy (i.e. purity) very close to many modern supervised approaches[1], whereas we only need ten labeled samples to remove the cluster-class permutation ambiguity.

## 5. Conclusions

We have presented a new clustering method based on nonnegative low-rank approximation with three major original contributions: (1) a novel decomposition approach for the approximating matrix derived from a two-step random walk; (2) a relaxed majorization-minimization algorithm for finding better approximating matrices; (3) a strategy that uses regularization with the Dirichlet prior as initialization. Experimental results showed that our method works robustly for various selected datasets and can improve clustering purity for large manifold datasets.

---

[1] see http://yann.lecun.com/exdb/mnist/

There are some other dimensions that affect clustering performance. Our practice indicates that initialization could play an important role because most current algorithms are only local optimizers. Using Dirichlet prior is only one way to smooth the objective function space. One could use other priors or regularization techniques to achieve better initializations.

Another dimension is the input graph. We have focused on the grouping procedure given that the similarities are precomputed. One should notice that better features or a better similarity measure can significantly improve clustering purity. Though we did not use AnchorGraph for the sake of including topic models in our comparison, it could be more beneficial to construct both approximated and approximating matrices by the same principle. This also suggests that clustering analysis could be performed in a deeper way using hierarchical pre-training. Detailed implementation should be investigated in the future.

## Acknowledgment

This work was financially supported by the Academy of Finland (Finnish Centre of Excellence in Computational Inference Research COIN, grant no. 251170; Zhirong Yang additionally by decision number 140398).

## References

Arora, R., Gupta, M., Kapila, A., and Fazel, M. Clustering by left-stochastic matrix factorization. In *International Conference on Machine Learning (ICML)*, pp. 761–768, 2011.

Asuncion, A., Welling, M., Smyth, P., and Teh, Y.-W. On smoothing and inference for topic models. In *Conference on Uncertainty in Artificial Intelligence (UAI)*, pp. 27–34, 2009.

Blei, D., Ng, A. Y., and Jordan, M. I. Latent dirichlet allocation. *Journal of Machine Learning Research*, 3: 993–1022, 2001.

Ding, C., Li, T., Peng, W., and Park, H. Orthogonal nonnegative matrix t-factorizations for clustering. In *International conference on Knowledge discovery and data mining (SIGKDD)*, pp. 126–135, 2006.

Ding, C., Li, T., and Jordan, M. I. Nonnegative matrix factorization for combinatorial optimization: Spectral clustering, graph matching, and clique finding. In *International Conference on Data Mining (ICDM)*, pp. 183–192, 2008.

Ding, C., Li, T., and Jordan, M. I. Convex and semi-nonnegative matrix factorizations. *IEEE Transactions on Pattern Analysis and Machine Intelligence*, 32(1):45–55, 2010.

Hein, M. and Bühler, T. An inverse power method for nonlinear eigenproblems with applications in 1-spectral clustering and sparse pca. In *Advances in Neural Information Processing Systems (NIPS)*, pp. 847–855, 2010.

Hofmann, T. Probabilistic latent semantic indexing. In *International Conference on Research and Development in Information Retrieval (SIGIR)*, pp. 50–57, 1999.

Liu, W., He, J., and Chang, S.-F. Large graph construction for scalable semi-supervised learning. In *International Conference on Machine Learning (ICML)*, pp. 679–686, 2010.

Mallat, S. Group invariant scattering. *Communications in Pure and Applied Mathematics*, 2012.

Minka, T. Estimating a dirichlet distribution, 2000.

Shi, J. and Malik, J. Normalized cuts and image segmentation. *IEEE Transactions on Pattern Analysis and Machine Intelligence*, 22(8):888–905, 2000.

Sinkkonen, J., Aukia, J., and Kaski, S. Component models for large networks. ArXiv e-prints, 2008.

Xu, W., Liu, X., and Gong, Y. Document clustering based on non-negative matrix factorization. In *International Conference on Research and Development in Informaion Retrieval (SIGIR)*, pp. 267–273, 2003.

Yang, Z. and Oja, E. Linear and nonlinear projective nonnegative matrix factorization. *IEEE Transaction on Neural Networks*, 21(5):734–749, 2010.

Yang, Z. and Oja, E. Unified development of multiplicative algorithms for linear and quadratic nonnegative matrix factorization. *IEEE Transactions on Neural Networks*, 22(12):1878–1891, 2011.

Zass, R. and Shashua, A. Doubly stochastic normalization for spectral clustering. In *Advances in Neural Information Processing Systems (NIPS)*, pp. 1569–1576, 2006.